# Feedback Without the Wait: Piloting a Generative AI Practice Platform in a Large Maths Class

Lili Chen[1,2], Gavin Buskes[1], Yuxin Ren[1] and Chin Tong Leong[1]
*The University of Melbourne[1] and Monash University[2]*
*Corresponding Author Email: lili.chen@monash.edu*

## ABSTRACT

### CONTEXT

Timely and specific feedback is one of the strongest influences on student learning, yet it is difficult to sustain in large electrical engineering classes where the ratio of students to demonstrators is high and a learner who is stuck may wait days to find out why an approach was wrong. Generative Artificial Intelligence (GenAI) offers a way to scale conversational feedback, but using it to grade assessed work raises trust and accountability concerns, and keeping a human in the loop to assure its judgements reintroduces the very delay that erodes the value of feedback. The result is a tension between the immediacy that makes feedback so impactful and the human oversight that makes it trustworthy.

### PURPOSE OR GOAL

In this work, we set out to resolve that tension in practice by designing and piloting a GenAI practice platform that delivers immediate, scaffolded feedback during self-directed practice. This relocates human oversight from real-time grading to the upfront verification of solutions. Our goal was to understand how students engaged with the tool, how they perceived the value and reliability of its feedback, and what lessons transfer to other engineering subjects.

### APPROACH OR METHODOLOGY/METHODS

The platform combines a demonstrator-verified question bank, AI-generated exam-style variants, and an iterative attempt–feedback–revise loop, and was deployed mid-semester in a core electrical engineering subject on probability. We drew on three sources of evidence: platform analytics for 95 registrants; an end-of-semester survey of 53 responses across five experiential constructs and four open-response items, analysed descriptively and thematically; along with first and second progress-test scores for a subset of engaged users, used to illustrate individual trajectories.

### ACTUAL OR ANTICIPATED OUTCOMES

Of 95 registrants, 34 attempted at least one question, generating 230 submissions. Users rated all five constructs above the neutral midpoint, where engagement and active learning rated best (3.97/5; 79% agreement) and lowest for preferring the system over the existing problem booklet (42%). Students who had used the platform were markedly more convinced than non-users, that immediate, imperfect AI feedback is more useful than delayed feedback from a demonstrator (79% versus 59%).

### CONCLUSIONS/RECOMMENDATIONS/SUMMARY

Decoupling feedback immediacy from solution verification and keeping the human in the loop upstream at answer verification, rather than at the point of grading, is a workable and transferable design pattern for trustworthy, scalable formative practice in engineering education. We offer practical recommendations on feed-forward prompt design, answer-input handling, and embedding such tools in the rhythm of a subject.

## Introduction

Feedback is among the most powerful influences on student achievement, but its effect depends heavily on being specific and timely (Hattie & Timperley, 2007; Shute, 2008). In large electrical engineering classes this is precisely where practice breaks down: as cohorts grow and the ratio of students to demonstrators rise, the depth and speed of formative feedback that teaching staff can sustain narrows (Pishchukhina & Allen, 2021). Students who get stuck on a problem often wait days, or even weeks, for a demonstrator to explain why their approach was wrong, by which time the productive moment for learning has passed.

Recent advances in generative artificial intelligence (GenAI) raise the possibility that scalable, conversational feedback could narrow this gap (Qadir, 2023). Early evidence suggests that GenAI can deliver formative feedback whose perceived quality approaches that of a teacher's feedback in some tasks (Chen et al., 2025), and that AI-generated hints can produce learning gains comparable to those from human-tutor hints in mathematics (Pardos & Bhandari, 2024). However, two obstacles stand in the way of simply handing assessment to an AI. First, students do not yet trust GenAI to grade their work; short informal interviews with students in our subject surfaced clear scepticism about AI marking. Second, where GenAI is used in assessment, good practice is to keep a human in the loop to check its judgements (The University of Melbourne, 2026), and that human review reintroduces the delay that undermines the value of feedback in the first place. The result is a tension between immediacy and trustworthiness.

In this paper we propose and instantiate a specific design methodology to address this tension: separating the generation of feedback from the verification of solutions. Rather than asking GenAI to grade live submissions, we position human markers as upfront verifiers of answers and use the model only to generate immediate, scaffolded feedback against those human-endorsed solutions. This pattern differs from conventional online quizzes, which often provide instant but generic correctness checks, and from fully automated GenAI graders, which can be immediate but lack the human oversight students and institutions still expect.

This paper reports how we applied this design methodology in practice. We built a platform for self-directed practice in which GenAI delivers immediate, scaffolded feedback against solutions that demonstrators have already verified. In effect, we repositioned the human from a real-time grader to an upfront verifier of answers, preserving immediacy where it matters pedagogically while retaining human oversight where students value it most. We describe the design, report what we observed when students were given access to the platform and distil practical lessons for other engineering educators. The paper offers four contributions: a design pattern that decouples feedback immediacy from solution verification; a workflow of integrating AI-generated and human-verified items in a single practice tool; reflection on uptake, perceptions and usability frictions from a real deployment; and a set of transferable recommendations for adapting the approach.

## Literature Review

### Formative Feedback and Active Learning

Formative feedback is among the most studied drivers of student achievement in higher education (Black & Wiliam, 2009; Shute, 2008). Hattie and Timperley's (2007) widely cited model argues that effective feedback addresses three questions for the learner: where am I going (feed-up), how am I going (feed-back), and where to next (feed-forward). The feed-forward component is consistently associated with stronger learning gains because it guides revision without merely revealing the answer. Shute (2008) similarly recommends that formative feedback be specific, elaborated, and delivered close in time to the learner's attempt.

These principles align with the broader case for active learning in STEM. Freeman et al.'s (2014) meta-analysis of 225 studies showed that active learning approaches reduced failure rates by 55% relative to traditional lectures across science, engineering, and mathematics. Active learning is most effective when paired with frequent low-stakes opportunities to attempt, fail, and revise — the precise pattern that iterative formative assessment supports (Black & Wiliam, 2009).

### Immediate Feedback and Computer-Based Practice

Computer-based formative assessment has long been used to scale feedback in large STEM cohorts. Van der Kleij, Feskens, and Eggen's (2015) meta-analysis reported a positive effect of feedback in computer-based learning environments, with elaborated feedback outperforming simple verification of correctness. Morris, Perry, and Wardle's (2021) systematic review found that immediate feedback on online quizzes tends to outperform delayed feedback in higher education, particularly when the feedback explains why an answer is incorrect. In an electrical and information engineering context, Pishchukhina and Allen (2021) showed that automated formative assessment with immediate feedback supported learning in a large cohort and reduced demonstrator workload, while noting that template-based feedback often lacked the diagnostic specificity students valued. The promise of GenAI is precisely the ability to generate that specificity at scale.

### Generative AI for Feedback

The release of capable large language models has prompted a rapid expansion of work on AI-generated feedback. Qadir (2023) outlined both the promise and the pitfalls of GenAI in engineering education, including concerns about hallucinated content and over-reliance. Kumar et al. (2024) examined how guiding students' use of large language models in a supported learning environment shaped their interaction patterns, performance, confidence, and trust. Pecuchova et al.(2026), when comparing GenAI, teacher, and no-feedback conditions across two engineering cohorts (N = 262), reported that students perceived GenAI feedback as beneficial and often comparable to teacher feedback, but that teacher feedback produced stronger performance gains on complex modelling tasks and that GenAI feedback was linguistically more repetitive. Cohn et al. (2026) showed that human-in-the-loop chain-of-thought prompting substantially improved LLM accuracy on formative-assessment scoring across science and engineering items.

Student trust in GenAI feedback is itself emergent: students often arrive with mixed prior beliefs and calibrate their trust through direct experience (Pecuchova et al., 2026). Motivation and trust are therefore plausible mediators of any observed effect of GenAI feedback on conceptual understanding, and both warrant explicit measurement. This is the crux of our design problem: human oversight is what makes GenAI judgements trustworthy in assessment, but human oversight delivered in real time does not scale, and human oversight delivered after the fact destroys immediacy. Our response was to relocate the human-in-the-loop step. Instead of reviewing each AI judgement as it is made, demonstrators verify the model solutions in advance; the GenAI then generates immediate feedback by reasoning against an answer a human has already endorsed. This keeps the verification students trust while removing the wait.

## The Platform: Design and Implementation

The platform was deployed in a core electrical engineering subject (Probability and Random Models) at the University of Melbourne during semester 1, 2026. The subject covers probability and random models and is taught to a large cohort through lectures and demonstrator-supported workshops, with practice traditionally supported by a problem booklet containing worked solutions. The architecture of GenAI platform that contains three elements. First, a curated bank of problem-booklet items drawn from existing course materials, with solutions verified by demonstrators. Second, exam-style questions generated by an AI agent to expand the pool of available practice beyond the fixed booklet, addressing a long-standing student complaint about insufficient practice questions. Third, an iterative feedback loop in which a student attempts a question, submits an answer, receives immediate GenAI feedback, and revises and resubmits.

#### Question Bank Construction and Knowledge Organisation

The platform maintains an original question bank that is kept separate from AI-generated variants. Existing course materials, including problem sheets, progress tests, and examination questions, were imported from LaTeX source documents using an automated processing pipeline. Question and solution files were parsed and converted into structured records before being loaded into the platform database. Questions are organised in a hierarchy of textbooks, knowledge points, original questions, and AI-generated variants. Each question stores its solution, question type, difficulty, and associated metadata. This structure supports topic-based practice, AI-assisted content generation, and knowledge-point-level analytics.

### AI Agent Architecture

The platform uses two specialised AI agents, implemented as separate LangGraph-based services (an open-source AI agent framework), to support question generation and formative feedback.

**Question Generation Agent:** To expand the available practice material, the question-generation agent creates new variants from existing questions and their associated knowledge points. Staff can trigger generation through the administrative interface by selecting a source question. The agent retrieves relevant question content and metadata from the structured question bank, together with previously generated variants to reduce duplication. It then generates a new question by modifying numerical parameters while preserving the underlying concepts being assessed, before producing updated answers and worked solutions. Generated content undergoes automated validation and is stored separately from the original question bank. Questions remain unavailable to students until reviewed and approved by staff.

**Feedback Generation Agent:** Student submissions are processed asynchronously by a dedicated LangGraph-based feedback agent. After a submission is recorded, the agent receives the question text, expected answer, question type, and, where available, the worked solution associated with the question. The agent returns structured outputs including correctness, score, and feedback text, which are stored and presented through the student interface. Rather than providing only a correctness judgement, the agent is designed to identify misconceptions, highlight missing reasoning steps, and suggest directions for improvement. Students may revise and resubmit answers, creating an iterative feedback cycle. Human oversight is incorporated throughout the content lifecycle. Original question-bank content is based on demonstrator-verified solutions, while AI-generated questions remain unavailable to students until reviewed and approved by staff. Students may also report issues with questions or feedback, which are reviewed and corrected through an administrative moderation workflow. Feedback prompts were designed around Hattie and Timperley's (2007) feed-forward principle: when a student's answer was incorrect, the system was instructed to indicate where the reasoning diverged and to guide the next step, rather than disclosing the final answer outright.

### Data collection

The platform is implemented by researchers. Access to the platform was voluntary and supplementary; it did not replace any existing tutorial or carry any assessment weight. Students engaging in the platform remained anonymous due to the requirements of using a fabricated profile. Students were encouraged to complete the survey at the end of semester. Participation in the survey was strictly voluntary.

## Findings and Reflections

The platform was released mid-semester, after the first progress test, as an optional practice resource. Students self-registered and used it at their own pace. Consistent with the reflective, practice-oriented aim of this paper, we drew on three sources of evidence. Platform analytics provided an objective record of uptake and engagement. An end-of-semester student survey captured perceptions across five experiential constructs aligned to the design intent. Progress-test scores for the first and second tests were available for a subset of users and are used below to illustrate individual trajectories. Of the 95 students (out of the 157 enrolled in the subject) who registered for the platform, 34 went on to attempt at least one question, altogether generating 230 submissions. Among these active users, engagement varied widely. The median student made three submissions, but the most active made 40, and ten students made ten or more.

## Linking Engagement to Progress: Three Illustrative Cases

To explore whether engagement related to attainment, we examined the first and second progress-test scores of the three most engaged users (each with more than 15 submissions) for whom both scores were available, expressing each as a percentile rank and a standardised score relative to the cohort (Table 1).

**Table 1. Engagement and progress-test trajectories for three illustrative users (anonymised).**

| Student | Platform engagement | Progress test (raw) | Standardised (z) | Percentile rank |
|---|---|---|---|---|
| Student A | 40 submissions | 28 → 38 (+10) | +0.77 → +1.98 (+1.21 SD) | 82nd → 99th (+17) |
| Student B | 25 submissions | 12 → 15 (+3) | −1.03 → −0.57 (+0.45 SD) | 13th → 32nd (+18) |
| Student C | 19 submissions | 21 → 29 (+8) | −0.02 → +0.98 (+1.00 SD) | 52nd → 82nd (+30) |

**Student A (the persistent, high-volume user).** Student A was the platform's heaviest user, submitting 40 attempts across five active days, and recorded the largest standardised gain of any user, moving from the 82nd to the 99th percentile of the cohort. The survey responses were candid: useful overall, although Student A initially found the interface difficult to navigate. Most tellingly for the formative intent, Student A described the system working them through problems they had abandoned, recounting that when they conceded they were stuck, it guided them step by step toward the solution.

**Student B (the low-starter who climbed).** Beginning near the bottom of the cohort (13th percentile; raw 12) and engaging heavily (25 submissions across three active days), Student B rose to the 32nd percentile, a more modest but real gain from a low base. Student B was also the most critical respondent, reporting that identical inputs sometimes returned different solutions, that the solution was occasionally incorrect, and that the system would benefit from more processing time. Even so, Student B valued being able to work through their own attempts, explaining that receiving feedback on a submitted solution helped them analyse their own reasoning. This balanced case, engagement and improvement alongside honest critique, strengthens rather than weakens the account.

**Student C (the focused, efficient user).** Student C used the platform more selectively, with 19 submissions across three active days, yet made the single largest ranking jump of the three, from the 52nd to the 82nd percentile, moving from mid-cohort into the top quintile. Student C did not complete the survey, so we present this case as a quantitative counterpoint to Student A, improvement through focused rather than high-volume use, without over-interpreting it.

Taken together, these three cases should be interpreted as illustrative trajectories: engagement was voluntary, the number of highly engaged students was small, and other supports or cohort-level factors may also have influenced progress-test performance. As such, these cases are best understood as practice-based vignettes that highlight patterns of interest and motivate more rigorous quasi-experimental analysis in future work.

## Perceptions across the five constructs

The survey was comprised of 11 survey items. We distinguish two respondent groups by cross-referencing the activity logs: platform users (n = 19), whose responses reflect direct experience, and non-users (n = 34), who completed the survey but had no recorded platform activity. The nine experience-based items are reported for users only; the two belief-based items, which probe general attitudes to immediate AI feedback, are compared across users and non-users. Survey responses were summarised descriptively, and open responses were analysed thematically following Braun and Clarke (2006). Thematic Analysis is a foundational qualitative research method to identify, analyse, and report themes within qualitative data, such as open response. Platform users' perceptions were positive across all five constructs, with every construct meaning above the neutral midpoint of three. Engagement and active learning were rated most highly (mean 3.97; 79% agreement), followed by usability and efficiency (3.82), conceptual understanding (3.68), and feedback effectiveness (3.54); while motivation and persistence was ranked the lowest (3.39; 55% agreement). However, the item-level picture (Figure 1) is more revealing than these averages. The single strongest item (Q1) was that students were actively thinking about the solution rather than guessing when using the platform (95% agreement), which speaks directly to the active-learning intent of the design. The weakest item (Q6) was a direct preference for the GenAI system over the existing problem booklet (42% agreement), where students were genuinely divided: a well-established resource with worked solutions remained the preferred option, and novelty alone was insufficient to displace it.

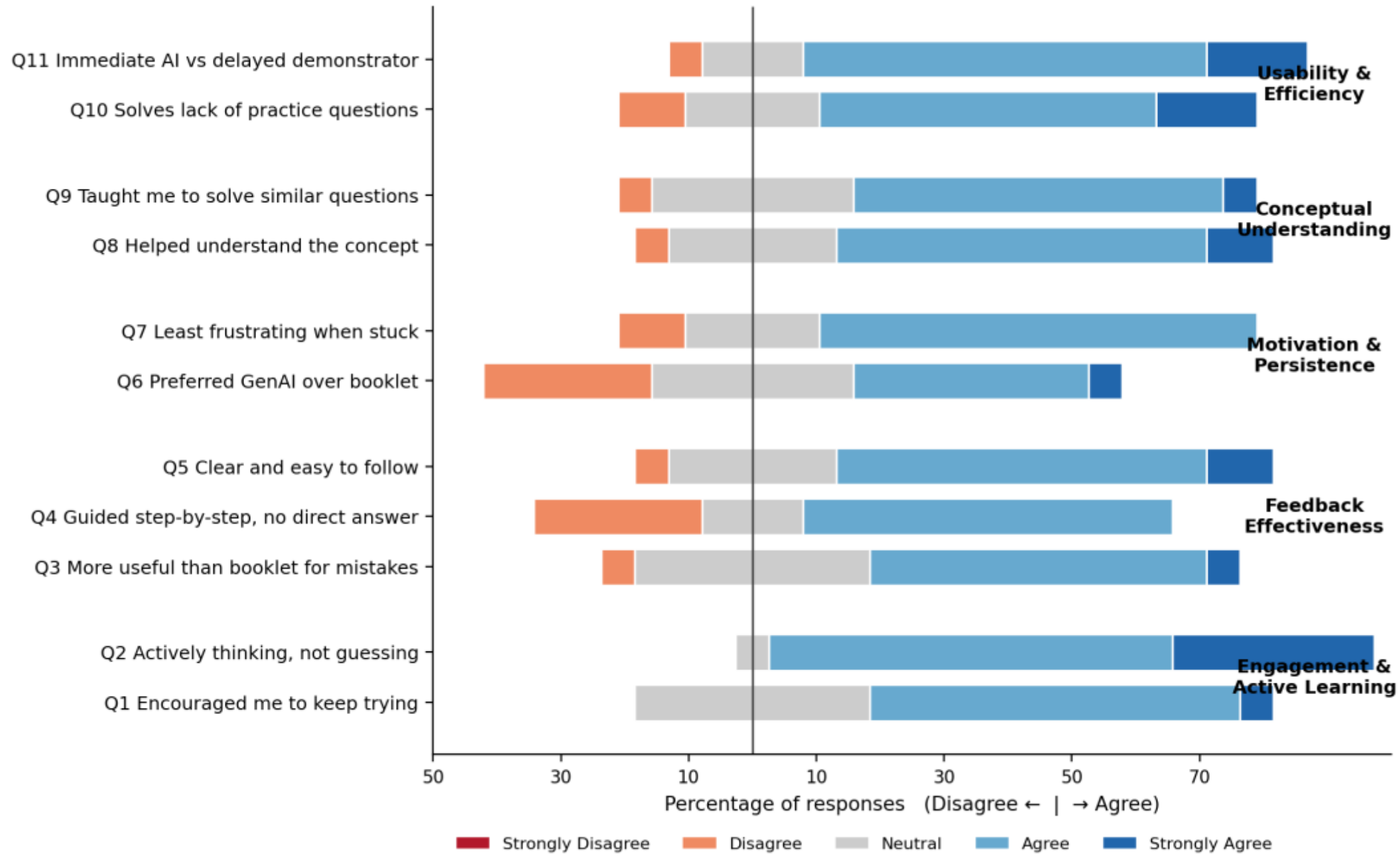

**Figure 1. Distribution of platform users' responses across the survey items, grouped by construct**

## Experience versus expectation

The two belief-based items allow a comparison between students who used the platform and those who did not (Figure 2). On whether the platform can address the shortage of practice questions, the two groups were similar (68% of users versus 62% of non-users agreed). This is reasonable as the platform only generates similar questions with slightly different parameters. The more striking result is on the belief that immediate AI feedback, even with potential errors, is more useful than delayed but higher-quality feedback from a demonstrator: 79% of users agreed, compared with only 59% of non-users. In other words, the students who had actually experienced the platform were considerably more convinced of the value of immediate AI feedback than those who had only imagined it. This is a useful corrective to a common worry that exposure to an imperfect AI would erode student trust. In our pilot the opposite held: direct experience appears to have strengthened, not weakened, confidence in the immediate-feedback value proposition. A plausible reading is that students traded a tolerance for small, visible errors against the benefit of never having to wait. This pattern is consistent with our qualitative finding of calibrated trust, and it supports the central design decision to position GenAI in immediate practice feedback rather than in grading.

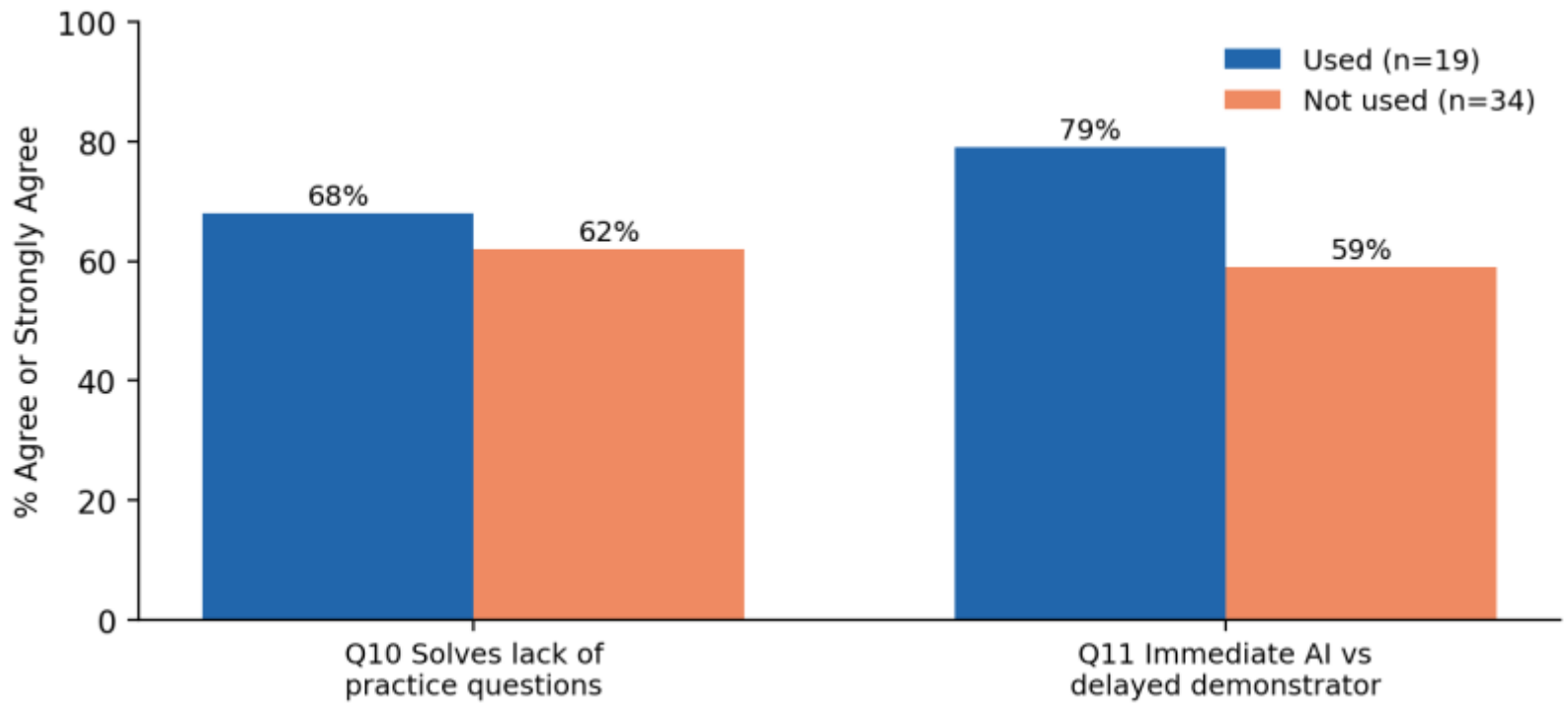

**Figure 2. Agreement on belief-based items: platform users versus non-users.**

### Qualitative Themes

Thematic analysis of the open responses surfaced four recurring themes, with representative student comments paraphrased in-line below.

**Immediacy as the core value:** Students valued getting feedback at the moment of difficulty, particularly where worked solutions alone had previously left them unsure why they were wrong. One wrote that they did not like leaving a question unresolved because it broke their thinking process, and that the platform helped them keep going and understand things step by step. Another noted that a past progress test had no published solutions, and that the platform gave them a way to obtain feedback and check their answers.

**Calibrated trust:** Students were realistic rather than naive about reliability. A representative comment acknowledged that the AI is not always like problem booklets with perfect solutions and sometimes makes mistakes, but found it helpful precisely because feedback arrived straight away. Students treating the platform as a complement to the booklet rather than a replacement. One student noted an appealing psychological benefit, that an AI does not know you, so you can ask what feel like basic questions without embarrassment. This nuanced stance supports our decision to position GenAI in practice rather than grading.

**Pinpointing the mistake:** A distinct theme was the value of feedback that engaged with a student’s own working and identified the specific step where it went wrong, rather than simply presenting the final answer. One student recalled entering a nearly correct answer and having the system pinpoint exactly where the mistake lay. Another, who had submitted a progress-test answer without understanding why it was wrong, found that the platform explained which key steps they had missed. The preview function was singled out: rather than showing only the final answer, it let a student see where their thinking had gone astray and identify the precise step they did not understand. Students wanted this depth applied consistently, asking for comprehensive guidance rather than simply being pointed to the answer. This is the feed-forward behaviour the design aimed for, and where it worked, students clearly recognised it.

**Progress and control:** Students wanted better ways to track which questions they had attempted, to select topics and question types, and in some cases to bypass feedback generation when already confident, noting the time and computational cost of generating individualised feedback for every submission. One found it difficult to keep track of which questions they had completed, noting that the system sometimes erased their previous answers, while another asked for adjustable levels of AI help.

## Implications For Practice and Transferability

Our experience suggests that the design methodology at the heart of this pilot, decoupling feedback immediacy from solution verification, transfers readily to other electrical engineering subjects that already maintain a bank of problems with worked solutions. The conditions that made it work, and the constraints others should anticipate, are as follows.

**Enabling conditions:** (a) An existing, demonstrator-verified solution set. The workflow depends on having human-endorsed solutions for the GenAI to reason against, therefore subjects with established problem booklets are well placed to adopt it. (b) Staff are comfortable with reviewing solutions of AI-generated questions. Verifying answers upfront is a bounded, familiar task for markers, unlike open-ended real-time oversight. (c) An institutional environment that permits GenAI for practice. Positioning the tool as practice rather than assessment eased trust and governance concerns for both students and staff.
**Constraints and boundary conditions:** (a) Voluntary, unweighted activities attract uneven engagement. Expect a substantial gap between registration and sustained use, unless as a continual activity of the subject. (b) Symbolic and notation-heavy answers need careful input handling. In mathematical engineering subjects, answer entry is a usability issue, not edge cases. (c) Generating feedback for every submission carries time and compute costs that students notice, therefore offering a fast path for confident learners may improve both experience and efficiency.
**Recommendations for educators:** (a) Keep the human in the loop at solution verification, not real-time grading, to preserve immediacy while retaining the oversight students trust; (b) Design feedback prompts explicitly around feed-forward: locate the error and guide the next steps rather than revealing the answer. Expect this to be the hardest behaviour to deliver consistently; (c) Invest early in symbolic-answer entry, preview, and algebraic-equivalence handling; pilot these with a small group before full release; (d) Provide clear progress tracking and topic or question-type selection so students can direct their own practice; (e) Embed the tool in the subject's workflow, for example by linking it to specific topics or low-stakes checkpoints, rather than releasing it as a standalone optional resource, to close the registration-to-use gap; (f) Be transparent with students that the AI can hallucinate and that solutions are human-verified; our experience suggests this supports calibrated trust rather than over-reliance.

## Conclusion

Faced with the tension between the immediacy that makes feedback valuable and the human oversight that makes it trustworthy, we built and piloted a GenAI practice platform that resolves the tension by moving human verification upstream of the feedback moment. In a large electrical engineering subject, students who used the platform engaged actively with immediate, scaffolded feedback and rated the experience positively. The pilot suggests that GenAI is most readily trusted, and most easily governed, when positioned as a practice companion working from human-verified solutions rather than as an autonomous grader. Future work will address the gaps in quantitative analysis. The present findings are based on a self-selected group of active users (n=19), small numbers of heavily engaged students, and observational survey and assessment data. Consequently, the findings should be interpreted as indicative rather than conclusive and cannot be assumed to generalise to the broader cohort. Building on these findings, a natural next step is a pre- and post-exposure survey that captures students' attitudes before and after they use the platform, allowing future work to trace how trust in immediate AI feedback develops through direct experience with the tool. Methodologically, building on the preliminary individual trajectories reported above, we plan a quasi-experimental analysis linking platform engagement to changes in student performance between the two progress tests, using baseline-adjusted comparisons and a larger, more representative sample.

## References

Black, P., & Wiliam, D. (2009). Developing the theory of formative assessment. *Educational Assessment, Evaluation and Accountability*,*21*(1),5–31. https://doi.org/10.1007/s11092-008-9068-5

Braun, V., & Clarke, V. (2006). Using thematic analysis in psychology. *Qualitative Research in Psychology*, *3*(2), 77–101. https://doi.org/10.1191/1478088706qp063oa

Chen, L., Chow, W. W. Y., Peng, S., Fan, B., & Bandara, S. (2025). Bridging qualitative rubrics and AI: A binary question framework for criterion-referenced grading in engineering. In *36th Australasian Association for Engineering Education Annual Conference (AAEE2025)* (pp. 1125-1134). Brisbane, QLD: Engineers Australia.

Cohn, C., T. S, A., Mohammed, N., & Biswas, G. (2026). CoTAL: Human-in-the-loop prompt engineering for generalizable formative assessment scoring and feedback. *arXiv preprint arXiv: 2504.02323*

Freeman, S., Eddy, S. L., McDonough, M., Smith, M. K., Okoroafor, N., Jordt, H., & Wenderoth, M. P. (2014). Active learning increases student performance in science, engineering, and mathematics. *Proceedings of the National Academy of Sciences*, *111*(23), 8410–8415. https://doi.org/10.1073/pnas.1319030111

Hattie, J., & Timperley, H. (2007). The power of feedback. *Review of Educational Research*, *77*(1), 81–112. https://doi.org/10.3102/003465430298487

Kumar, H., Musabirov, I., Reza, M., Shi, J., Wang, X., Williams, J. J., Kuzminykh, A., & Liut, M. (2024). Guiding students in using LLMs in supported learning environments: Effects on interaction dynamics, learner performance, confidence, and trust. *Proceedings of the ACM on Human-Computer Interaction*, *8*(CSCW2), 1–30. https://doi.org/10.1145/3687038

Morris, R., Perry, T., & Wardle, L. (2021). Formative assessment and feedback for learning in higher education: A systematic review. *Review of Education*, *9*(3), e3292. https://doi.org/10.1002/rev3.3292

Pardos, Z. A., & Bhandari, S. (2024). ChatGPT-generated help produces learning gains equivalent to human tutor-authored help on mathematics skills. *PLOS ONE*, *19*(5), e0304013. https://doi.org/10.1371/journal.pone.0304013

Pecuchova, J., Benko, L., & Drlik, M. (2026). Reimagining feedback through generative AI in engineering education. *Computers and Education: Artificial Intelligence*, *10*, 100574. https://doi.org/10.1016/j.caeai.2026.100574

Pishchukhina, O., & Allen, A. (2021). Supporting learning in large classes: Online formative assessment and automated feedback. *2021 30th Annual Conference of the European Association for Education in Electrical and Information Engineering (EAEEIE)*, 1–4. https://doi.org/10.1109/EAEEIE50507.2021.9530953

Qadir, J. (2023). Engineering education in the era of ChatGPT: Promise and pitfalls of generative AI for education. *2023 IEEE Global Engineering Education Conference (EDUCON)*, 1–9. https://doi.org/10.1109/EDUCON54358.2023.10125121

Shute, V. J. (2008). Focus on formative feedback. *Review of Educational Research*, *78*(1), 153–189. https://doi.org/10.3102/0034654307313795

The University of Melbourne. (2026, March 4). *Using AI for student assessment and feedback*. Generative AI. https://www.unimelb.edu.au/ai/home/staff/teaching-and-learning/menu-items/assessment

Van Der Kleij, F. M., Feskens, R. C. W., & Eggen, T. J. H. M. (2015). Effects of feedback in a computer-based learning environment on students' learning outcomes: A meta-analysis. *Review of Educational Research*, *85*(4), 475–511. https://doi.org/10.3102/0034654314564881

## Ethics statement

This study was approved under the Faculty's Teaching and Learning Laboratory Teaching Evaluation Protocol of the University of Melbourne (Approval No. 29248).

## Copyright statement